%% file: paper.tex
\documentclass{article}

\pdfoutput=1

\usepackage{microtype}
\usepackage{graphicx}
\usepackage{subfigure}
\usepackage{booktabs} 

\usepackage{hyperref}

\usepackage[accepted]{icml2024}

\usepackage{amsmath}
\usepackage{amssymb}
\usepackage{mathtools}
\usepackage{amsthm}
\usepackage{multirow}
\usepackage{makecell}
\usepackage{caption}
\usepackage{titlesec}

\usepackage[capitalize,noabbrev]{cleveref}

\theoremstyle{plain}

\theoremstyle{definition}

\theoremstyle{remark}

\usepackage[textsize=tiny]{todonotes}

\usepackage{xspace}
\makeatletter
\DeclareRobustCommand\onedot{\futurelet\@let@token\@onedot}
\def\@onedot{\ifx\@let@token.\else.\null\fi\xspace}

\def\ie{\emph{i.e}\onedot}

\makeatother

\icmltitlerunning{FVMD: A Metric for Evaluating Motion Consistency in Videos}

\begin{document}

\twocolumn[
\icmltitle{Fr\'echet Video Motion Distance: \\ A Metric for Evaluating Motion Consistency in Videos}




\begin{icmlauthorlist}
\icmlauthor{Jiahe Liu}{ubc,vector}
\icmlauthor{Youran Qu}{pku}
\icmlauthor{Qi Yan}{ubc,vector}
\icmlauthor{Xiaohui Zeng}{uoft}
\icmlauthor{Lele Wang}{ubc}
\icmlauthor{Renjie Liao}{ubc,vector,cifar}

\end{icmlauthorlist}

\icmlaffiliation{ubc}{University of British Columbia}
\icmlaffiliation{pku}{Peking University}
\icmlaffiliation{uoft}{University of Toronto}
\icmlaffiliation{vector}{Vector Institute for AI}
\icmlaffiliation{cifar}{Canada CIFAR AI Chair}


\icmlcorrespondingauthor{Renjie Liao}{rjliao@ece.ubc.ca}

\icmlkeywords{Machine Learning, ICML}

\vskip 0.3in
]



\printAffiliationsAndNotice{}  



\input{sec/0_figure1}

\input{sec/0_abstract} 
\input{sec/1_intro}
\input{sec/2_related_work}
\input{sec/3_method}
\input{sec/4_experiment}
\input{sec/5_conclusion}

\input{sec/acknowledgement}

\bibliography{paper}
\bibliographystyle{icml2024}

\input{sec/appendix}

\end{document}

%% file: sec/0_figure1.tex
\begin{figure*}[!ht]
    \centering
    \vspace{-0.3cm}
    \begin{minipage}[b]{0.9\textwidth}
        \centering
        \includegraphics[width=\linewidth]{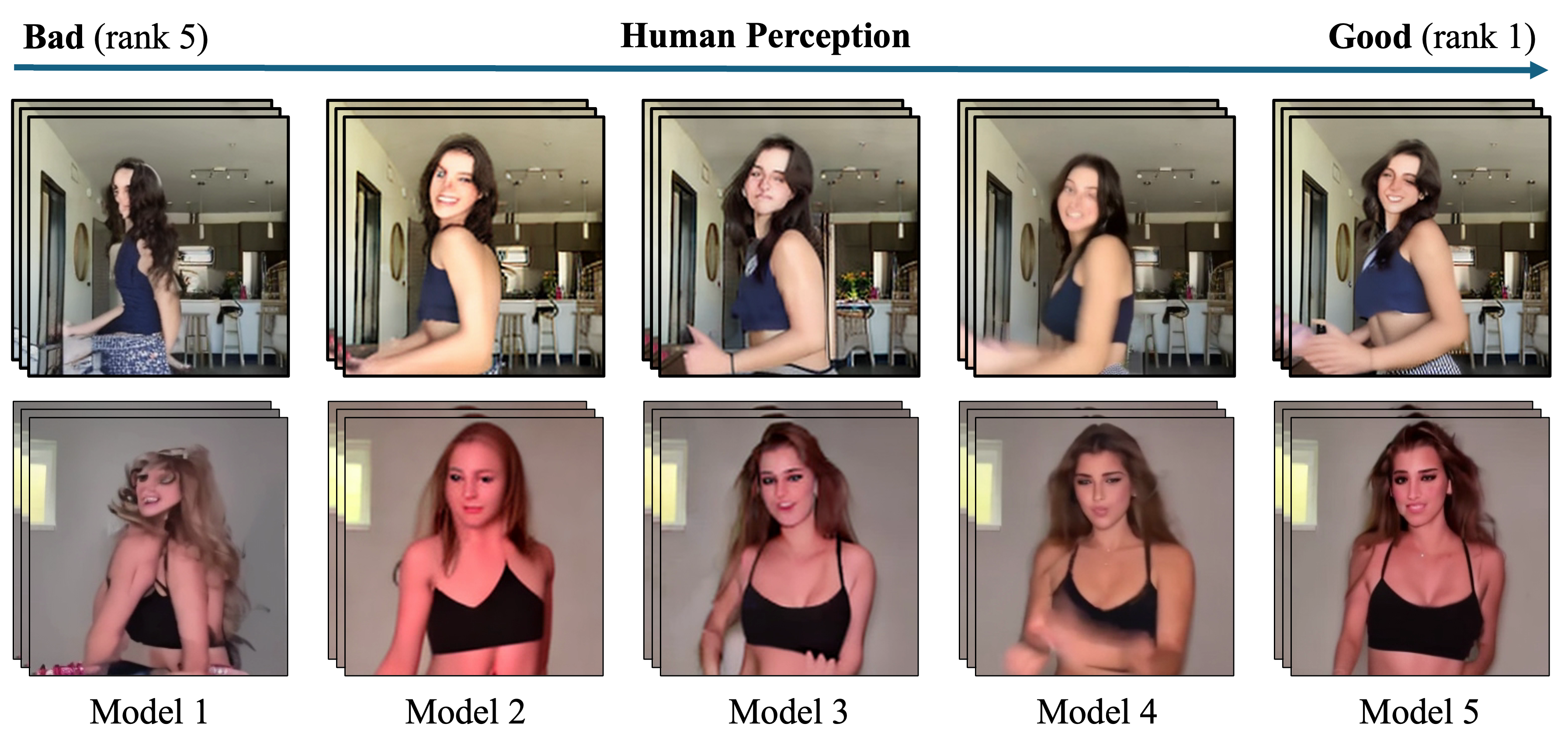}
        \label{fig:your_figure_label}
    \end{minipage}\\[-2ex] 
    \begin{minipage}[b]{0.99\textwidth}
        \centering
        \begin{tabular}{l|ccccc|c}
                \toprule
                \multirow{2}{*}{\textbf{Metrics}} & \multirow{2}{*}{\textbf{Model 1}} & \multirow{2}{*}{\textbf{Model 2}} & \multirow{2}{*}{\textbf{Model 3}} & \multirow{2}{*}{\textbf{Model 4}} & \multirow{2}{*}{\textbf{Model 5}} &  \textbf{Correlation $\uparrow$} \\
                & & & & & & \textbf{(w.r.t human)}\\
                \midrule
                FVMD$\downarrow$     & 7765.91/5 & 3178.80/4 & 2376.00/3 & 1677.84/2 & 926.55/1 & \textbf{0.8469} \\
                FVD$\downarrow$      & 405.26/4  & 468.50/5  & 247.37/2  & 358.17/3  & 147.90/1 & 0.6708\\
                FID-VID$\downarrow$  & 73.20/3   & 79.35/4   & 63.15/2   & 89.57/5   & 18.94/1 & 0.3402 \\
                VBench$\uparrow$   & 0.7430/5  & 0.7556/4  & 0.7841/2  & 0.7711/3  & 0.8244/1 & 0.7573\\
                \bottomrule
        \end{tabular}
        \caption{
        \textbf{Comparison of the fidelity of different video evaluation metrics.}
        \textbf{Top}: we present videos generated by various models trained on the TikTok dataset~\cite{jafarian2022self}, ranked according to the human ratings in the user study. 
        \textbf{Bottom}: we show quantitative scores and relative ranking given by our FVMD and other widely-used metrics, including FVD~\cite{unterthiner2018towards}, FID-VID~\cite{balaji2019conditional}, and VBench~\cite{huang2023vbench}. The correlations are computed using the Pearson correlation coefficient with human scores (detailed in~\cref{subsec:human}).
        Our FVMD achieves the best correlation with human judgment among all the metrics and clearly distinguishes video samples of different quality.
        }
        \label{tab:your_table_label}
    \end{minipage}
    \vspace{-1em}
\end{figure*}

%% file: sec/0_abstract.tex
\begin{abstract}
Significant advancements have been made in video generative models recently. 
Unlike image generation, video generation presents greater challenges, requiring not only generating high-quality frames but also ensuring temporal consistency across these frames. 
Despite the impressive progress, research on metrics for evaluating the quality of generated videos, especially concerning temporal and motion consistency, remains underexplored. 
To bridge this research gap, we propose \emph{Fr\'echet Video Motion Distance (FVMD)} metric, which focuses on evaluating motion consistency in video generation. 
Specifically, we design explicit motion features based on key point tracking, and then measure the similarity between these features via the Fr\'echet distance.
We conduct sensitivity analysis by injecting noise into real videos to verify the effectiveness of FVMD.
Further, we carry out a large-scale human study, demonstrating that our metric effectively detects temporal noise and aligns better with human perceptions of generated video quality than existing metrics.
Additionally, our motion features can consistently improve the performance of Video Quality Assessment (VQA) models, indicating that our approach is also applicable to unary video quality evaluation.
Code is available at \url{https://github.com/ljh0v0/FMD-frechet-motion-distance}.
\end{abstract}
\vspace{-2em}

%% file: sec/1_intro.tex
\section{Introduction}
\label{sec:intro}

Recently, diffusion models have demonstrated remarkable capabilities in high-quality image generation~\cite{sohl2015deep, song2019generative, ho2020denoising}. This advancement has been extended to the video domain, giving rise to text-to-video diffusion models~\cite{ho2022video,singer2022make,ho2022imagen,zhou2022magicvideo,he2022latent}.  Compared to prior works, state-of-the-art diffusion-based video generation models, such as Sora~\cite{videoworldsimulators2024}, not only aim to generate visually impressive videos but also focus on challenges involving diverse and complex motions, including intricate human dance videos, thrilling fight scenes in movies and sophisticated camera movements. 
In this case, measuring the motion consistency of these generated videos emerges as a significant research question.

Despite the rapid development of video generation models, research on evaluation metrics for video generation remains insufficient. Currently, FID-VID~\cite{balaji2019conditional} and FVD~\cite{unterthiner2018towards} are widely used to measure the quality of generated videos. FID-VID assesses the visual quality of generated videos by comparing synthesized video frames to real reference video frames, neglecting the video motion quality. In contrast, FVD introduces an evaluation of temporal coherence by extracting video features using a pre-trained action recognition model, Inflated 3D Convnet (I3D)~\cite{carreira2017quo}. Recently, VBench provides a comprehensive 16-dimensional evaluation suite for text-to-video generative models~\cite{huang2023vbench}. Nevertheless, the evaluation protocols in VBench for temporal consistency, such as temporal flickering and motion smoothness, tend to award videos with smooth or even static movement, while overlooking high-quality videos with intensive motion, such as dancing and sports videos. Consequently, there is currently no metric specifically designed to evaluate the complex motion patterns in generated videos. This oversight is particularly evident in tasks like motion guided video generation. In these tasks, FID-VID and FVD can only measure whether the appearance of the generated video is consistent with the reference video, but not whether the motion matches the target motion. For VBench, both high-quality and low-quality videos are favored in terms of dynamic degree and penalized for temporal flickering and motion smoothness. This is because the ground-truth videos exhibit intense movements, leading to VBench giving inconsistent assessments compared to human judgement.

To address this research gap, we propose the \emph{Fr\'echet Video Motion Distance (FVMD)}, a novel metric that focuses on the motion consistency of video generation. Our main idea is to measure temporal motion consistency based on the patterns of velocity and acceleration in video movements, as motions conforming to real physical laws should not exhibit sudden changes in acceleration. Specifically, we extract the motion trajectory of key points in videos using a pre-trained point tracking model, PIPs++~\cite{zheng2023pointodyssey}, and compute the velocity and acceleration for all key points across video frames. We then obtain the motion features based on the statistics of the velocity and acceleration vectors. Finally, we measure the similarity between the motion features of generated videos and ground-truth videos using Fr\'echet distance~\citep{dowson1982frechet}. 
Our key contributions are as follows: 1) We propose the \emph{Fr\'echet Video Motion Distance (FVMD)}, a novel metric for video generation focusing on motion consistency. 
2) We conduct extensive experiments to evaluate our metric, including sensitivity analysis and human studies, demonstrating our metric is effective in capturing temporal noise and aligns better with human perceptions of video quality than existing metrics.
3) When applied to the Video Quality Assessment (VQA) task, our proposed motion feature leads to consistently improved performances, suggesting the universality of our method and its potential for generic video evaluation tasks.

%% file: sec/2_related_work.tex
\section{Related Work}
\label{sec:relatedwork}

\noindent\textbf{Video Generation.}
Video generation has long been a challenging and essential area of research. Previous studies have explored various model architectures to tackle this task, such as recurrent neural networks (RNNs)~\cite{babaeizadeh2017stochastic,castrejon2019improved,denton2018stochastic,franceschi2020stochastic,lee2018stochastic}, autoregressive transformers~\cite{yan2021videogpt,wu2022nuwa,hong2022cogvideo,ge2022long,villegas2022phenaki}, normalizing flows~\cite{blattmann2021ipoke,dorkenwald2021stochastic}, and generative adversarial networks (GANs)~\cite{vondrick2016generating,saito2017temporal,wang2019few,skorokhodov2022stylegan,voleti2022mcvd}.

Recently, diffusion models have proven to be powerful tools for image generation tasks and have since been applied to the video field, starting with unconditional generation. VDM~\cite{ho2022video} presents the first results on video generation based on diffusion models by inserting additional temporal attention blocks into the original 2D U-Net model. Make-A-Video~\cite{singer2022make} and Imagen Video~\cite{ho2022imagen} both propose cascaded spatial-temporal up-sampling pipelines to generate long videos with high resolution. LVDM~\cite{he2022latent} follows the latent diffusion paradigm, lightening and accelerating the video diffusion model by adapting it to the low-dimensional 3D latent space. 

Beyond unconditional video generation, significant advancements have been made in video generation conditioned on other modalities, inspired by the success of conditional models like ControlNet in the image domain~\cite{zhang2023adding}. One notable area is pose-guided video generation, where the goal is to generate videos that adhere to a specified pose sequence, providing control over the motion in the video. Disco~\cite{wang2023disco} leverages ControlNet and proposes a novel model architecture with disentangled control to improve the compositionality of human dance synthesis. Animate Anyone~\cite{hu2023animate} and Magic Animate~\cite{xu2023magicanimate} improve on Disco by adding a motion module to maintain temporal consistency.

\noindent\textbf{Video Evaluation Metrics.}
Quantitative evaluation metrics can be categorized into frame-level and video-level metrics. Commonly employed frame-level metrics include the Fr\'echet Inception Distance (FID-VID)~\cite{balaji2019conditional}, Peak Signal-to-Noise Ratio (PSNR)~\cite{wang2004image}, Structural Similarity Index Measure (SSIM)~\cite{wang2004image}, and CLIP similarity~\cite{radford2021learning}. FID-VID assesses the generated frames by extracting image features using a pre-trained image classification model, Inception v3~\cite{szegedy2016rethinking}, fitting a Gaussian distribution, and measuring the Fr\'echet Distance with the ground-truth frames. PSNR is a coefficient representing the ratio between the peak signal and Mean Squared Error (MSE). SSIM is a pixel-level metric that evaluates the luminance, contrast, and structure between generated and reference frames. CLIP similarity measures the alignment between image and text features obtained by the pre-trained CLIP model.

Compared to frame-level metrics, which focus solely on the quality of individual frames, video-level metrics capture both the temporal coherence of a video and the quality of each frame. The Fr\'echet Video Distance (FVD)~\cite{unterthiner2018towards} is a widely used video-level metric. It follows the assumptions of FID and replaces the image classification model with a pre-trained Inflated 3D Convnet (I3D)~\cite{carreira2017quo}. Similar to FVD, Kernel Video Distance (KVD)~\cite{unterthiner2018towards} employs the same I3D model but utilizes the Maximum Mean Discrepancy (MMD) to measure similarity. Video Inception Score (IS)~\cite{saito2020train} calculates an inception score based on 3D ConvNets (C3D)~\cite{tran2015learning}. 

Recently, VBench~\cite{huang2023vbench} has been proposed to provide a comprehensive benchmark suite that dissects video quality into hierarchical dimensions, each with tailored prompts and evaluation protocols. The motion-related metrics include temporal flickering, motion smoothness, and dynamic degree. Temporal flickering detects video inconsistency by computing the Mean Absolute Error (MAE) across frames. Motion smoothness evaluates the MAE between the generated frames and synthetic frames using frame interpolation. Since the first two dimensions tend to favor static videos, dynamic degree, which measures the extent of motion in the video, is proposed to counter this effect. 
However, VBench has considerable limitations, particularly for videos involving intensive motion. For instance, in the task of generating TikTok dancing videos, VBench does not clearly distinguish between high-quality and low-quality samples (see~\cref{subsec:human} for details). This is because both types of videos exhibit a high degree of dynamics and large differences between adjacent frames due to the large amplitude of movements. In contrast, even in the presence of intensive motion, our FVMD prefers high-quality videos over low-quality ones, resulting in more accurate scoring.

%% file: sec/3_method.tex
\section{Method}
\input{sec/0_figure2}
We propose the Fr\'echet Video Motion Distance (FVMD), a new video generation metric that measures the discrepancy in motion features between generated videos and ground-truth videos. The overall pipeline is illustrated in~\cref{fig:pipeline}.

\subsection{Motion Feature Extraction}
\par\noindent\textbf{Video Key Point Tracking.}
To construct video motion features, we first track key point trajectories across the video sequence. 
We utilize the PIPs++ model~\cite{zheng2023pointodyssey}, a state-of-the-art key point tracking approach built upon the particle video method~\cite{sand2008particle}, for this purpose. 
The selection of PIPs++ is motivated by two key benefits: 
1) PIPs++ predicts plausible positions for missing objects in the presence of occlusions, out-of-bounds movements, or difficult lighting conditions. 
This capability is essential for obtaining a consistent and robust motion trajectory, especially in generated videos where objects may become distorted, blurred, or abruptly vanish. 
2) PIPs++ estimates the trajectory of every tracking target independently, allowing computation to be shared between particles within a video, which enhances the speed of inference. 

For a set of $m$ generated videos, denoted as $\{{X^{(i)}}\}_{i=1}^m$, the tracking process starts by truncating longer videos into segments of $F$ frames with an overlap stride of $s$. Subsequently, we query $N$ target points in a grid shape on the initial frames. PIPs++ is then engaged to estimate $N$ trajectories, denoted as $\hat{Y} \in \mathbb{R}^{F \times N \times 2}$, for these key points, where each trajectory has a coordinate dimension of 2.

\par\noindent\textbf{Key Point Velocity and Acceleration Fields.}
To obtain representations of motion patterns, we compute the velocity field and acceleration field for each frame within the videos. The temporal inconsistencies in generated videos, such as unnatural changes in object position or posture, sudden deformations or blurring of objects, and jerky movements, can be reflected in disordered key point trajectories, resulting in abrupt changes in the velocity and acceleration of key points. Therefore, the patterns of velocity and acceleration changes over time can effectively indicate whether a video is temporally consistent.

The velocity field $\hat{V} \in \mathbb{R}^{F \times N \times 2}$ measures the first-order difference in key point positions between consecutive frames.
To have the same shape as the trajectories $\hat{Y}$, we pad the initial frames in $\hat{V}$ with a zero-frame. 
The velocity field $\hat{V}$ for a $F$-frames video segmentation is computed by:
\begin{equation}
    \hat{V} = \texttt{concat}(\boldsymbol{0}_{N\times 2}, \hat{Y}_{2:F} -  \hat{Y}_{1:F-1}) \in \mathbb{R}^{F \times N \times 2},
    \label{eq:vel_field}
\end{equation}
where $\boldsymbol{0}_{N\times 2}$ is the zero-padding frame whose subscript indicates its shape.
We use $\hat{Y}_{i:j}$ (or $\hat{V}_{i:j}$) to denote the range of frames from the $i$-th to the $j$-th inclusively.

Similarly, the acceleration field $\hat{A} \in \mathbb{R}^{F \times N \times 2}$ can be calculated by the first-order difference between the velocity fields in two consecutive frames.
Likewise, we pad the first frame of $\hat{A}$ to maintain the same shape as the input:
\begin{equation}
    \hat{A} = \texttt{concat}(\boldsymbol{0}_{N\times 2}, \hat{V}_{2:F} -  \hat{V}_{1:F-1}) \in \mathbb{R}^{F \times N \times 2},
    \label{eq:accl_field}
\end{equation}
where the subscripts align with those in~\cref{eq:vel_field}.

\par\noindent\textbf{Motion Feature.}
To obtain compact motion features, we further process the velocity and acceleration fields into spatial and temporal statistical histograms.
First, we compute the magnitude and angle for each key point in the velocity and acceleration vector fields respectively. Let $\rho(U)$ and $\phi(U)$ denote the magnitude and angle of a vector field $U$, where $U \in \mathbb{R}^{F \times N \times 2}$ and $U$ can be either $\hat{V}$ or $\hat{A}$. For each frame indexed by $i \in [F]$ and each point indexed by $j \in [N]$, we calculate the magnitude using the $l_2$ norm and the angle using the inverse hyperbolic tangent $\tanh^{-1}$. The equations are defined as follows:
\begin{align}
\rho(U)_{i, j} &= \sqrt{U_{i,j,1}^2 + U_{i,j,2}^2}, \forall i \in [F], j \in [N], \\
\phi(U)_{i, j} &= \left| \tanh^{-1}\left(\frac{U_{i, j,1}}{U_{i, j,2}}\right) \right|, \forall i \in [F], j \in [N].
\end{align}
Next, the magnitudes $\rho$ are clipped to a range of $[0,255]$. Given that most vector fields have small magnitudes, a base-2 logarithmic transformation is applied for normalization. The magnitudes are then quantized to the nearest integer, resulting in nine discrete bins in range of $[0, 8]$.
We also quantize the angle representations $\phi$ into 8 bins, with each bin encompassing an angle range of 45 degrees. 

\begin{figure*}[!ht]
    \centering
    \vspace{-0.30cm}\includegraphics[width=0.33\textwidth]{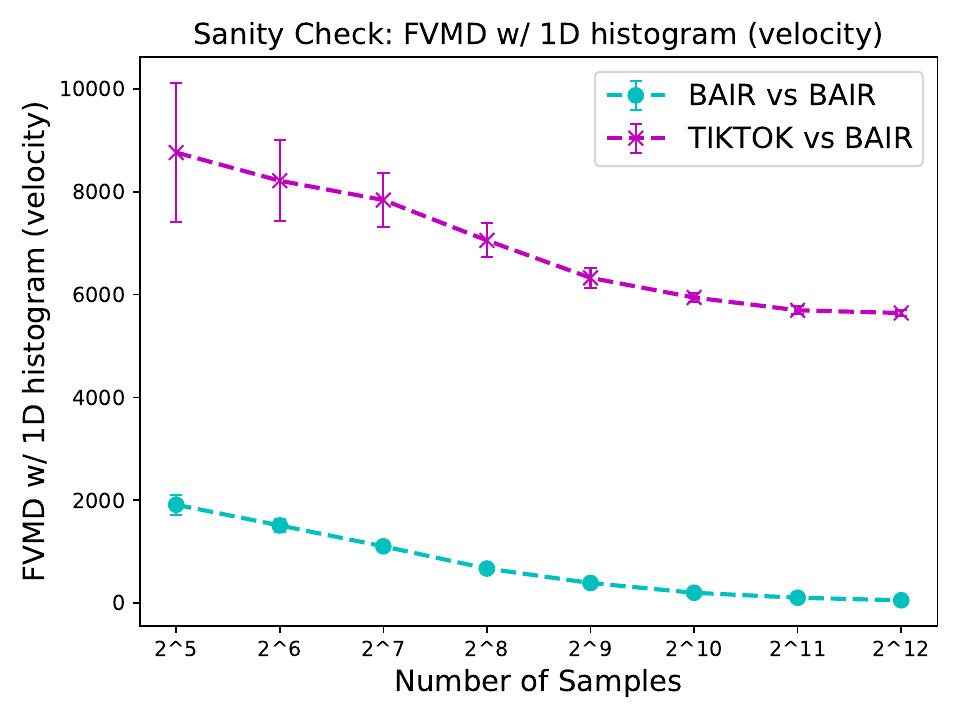} 
    \includegraphics[width=0.33\textwidth]{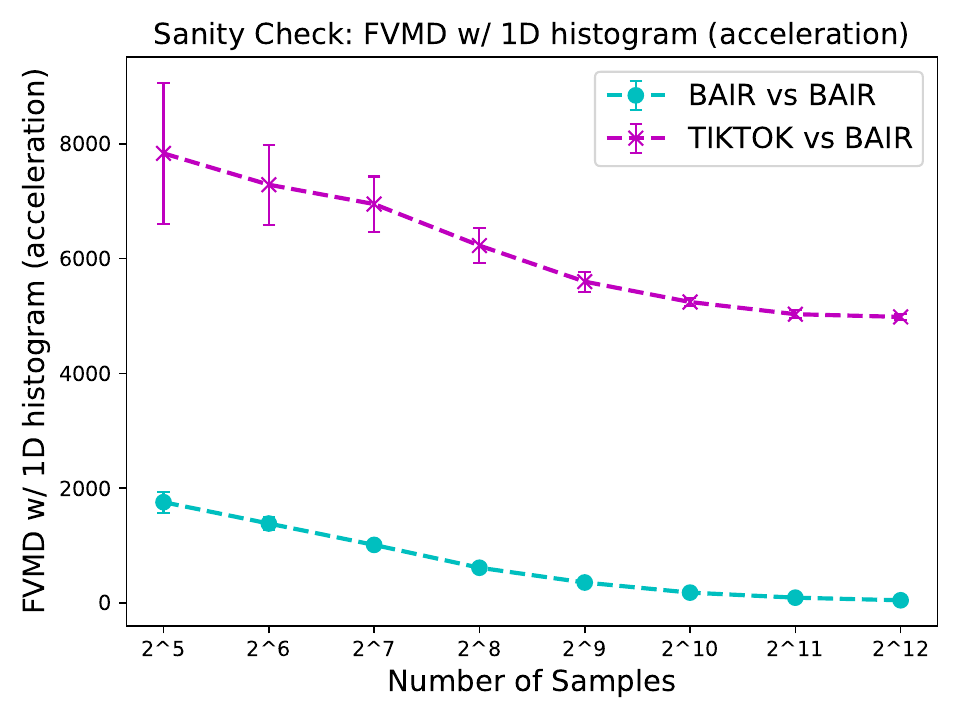} 
    \includegraphics[width=0.33\textwidth]{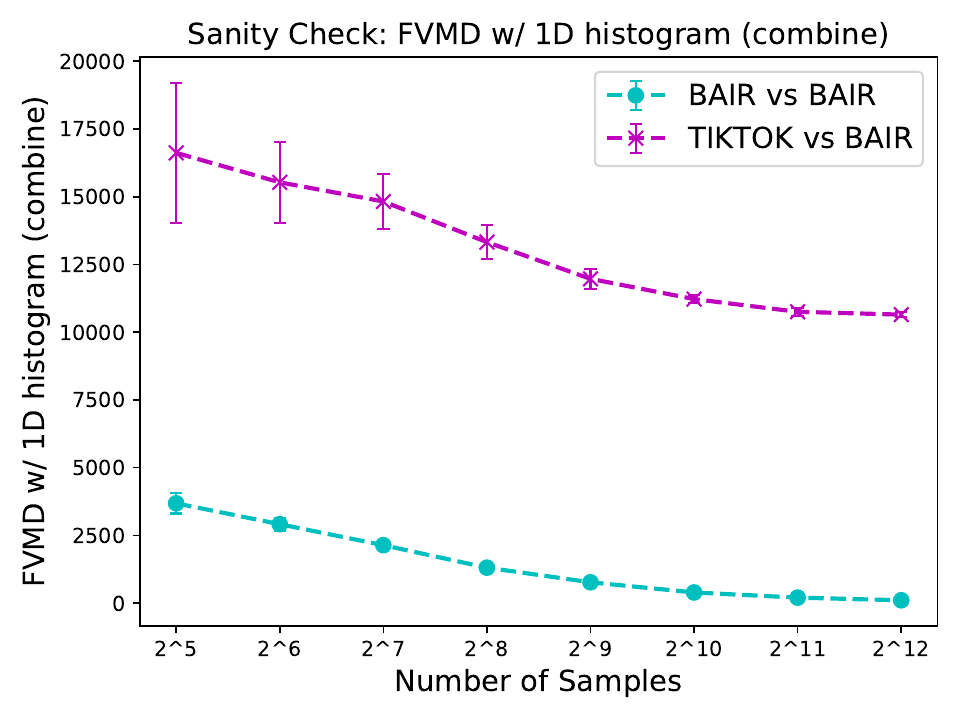}
    \vspace{-0.80cm}
    \caption{
    \textbf{Sanity check experiments.}
    We use dense 1D histograms based on velocity, acceleration, and their concatenated combination to construct FVMD metrics. As sample size increases, same-dataset discrepancies (BAIR vs BAIR) converge to zero, while cross-dataset discrepancies (TIKTOK vs BAIR) remain large, verifying the soundness of our FVMD metric.
    }
    \label{fig:sanity_check}
    \vspace{-1.3em}
\end{figure*}

We employ two methods for calculating statistical histograms of the quantized magnitudes and angles. The first utilizes a quantized \textbf{2D histogram}. We divide the $F$-frame video segments into smaller volumes of size $f \times k \times k$, where $f$ is the number of frames and $k$ the spatial dimensions of each volume. In each volume, we aggregate all vectors to form a 2D histogram, with $x$ and $y$ coordinates corresponding to magnitudes and angles, respectively. These 2D histograms are then concatenated and flattened into a vector, forming the motion feature for the respective video segment. The shape of the quantized 2D histogram is $\frac{F}{f} \times \frac{\sqrt{N}}{k} \times \frac{\sqrt{N}}{k} \times 72$, where the number 72 is derived from 8 discrete bins for angle and 9 bins for magnitude.

Inspired by the HOG (Histogram of Oriented Gradients) approach~\cite{dalal2005histograms}, which counts occurrences of gradient orientation in localized portions of an image, we compile a 2D histogram into a dense \textbf{1D histogram} focused on the angle dimension.  Similarly, we divide each video segmentation into small $f \times k \times k$ volumes. Our goal is to create a 1D histogram with 8 bins, each corresponding to a range of quantized angles. Within each volume, magnitudes are summed directly into the appropriate angle bin, resulting in an 8-point histogram per volume. By combining these histograms from all volumes, we create the final motion feature, shaped as $\frac{F}{f} \times \frac{\sqrt{N}}{k} \times \frac{\sqrt{N}}{k} \times 8$.

We apply these two histogram counting methods to separately build the motion features for both velocity and acceleration fields. Additionally, we explore concatenating features from these fields to form a combined motion feature, which can then be used to compute similarity.

\subsection{Fr\'echet Video Motion Distance}
After extracting motion features from video segments of generated and ground-truth video sets, we measure their similarity using the Fr\'echet distance~\citep{dowson1982frechet}, which we have named the \textbf{Fr\'echet Video Motion Distance (FVMD)}:
\begin{equation}
    \scalebox{0.85}{$
    d_F(P_{\text{data}}, P_{\text{gen}}) = \left(\inf_{\gamma \in \Gamma (P_{\text{data}}, P_{\text{gen}})} \int \lVert x-y \rVert^2_2 d\gamma(x, y) \right)^{\frac{1}{2}},
    $}
\end{equation}
where the $P_{\text{gen}}$ denotes the distribution of motion features for generated videos, $P_{\text{data}}$ denotes the distribution of motion features for ground-truth videos, and $\Gamma (P_{\text{data}}, P_{\text{gen}})$ is the set of all couplings of $P_{\text{gen}}$ and $P_{\text{data}}$. 
However, $P_{\text{data}}$ and $P_{\text{gen}}$ are normally intractable and there is no analytic expression for the Fr\'echet distance between two arbitrary distributions. 
Hence, we follow FID~\cite{balaji2019conditional} to approximate the distributions with multivariate Gaussians. 
In this case, the Fr\'echet distance has a closed-form solution:
\begin{equation}
    \scalebox{0.88}{$
    d_F = \lVert\mu_{\text{data}}-\mu_{\text{gen}}\rVert_2^2 + \mathrm{tr}\left(\Sigma_{\text{data}} + \Sigma_{\text{gen}} -2(\Sigma_{\text{data}}\Sigma_{\text{gen}})^{\frac{1}{2}}\right),
    $}
\end{equation}
where the $\mu_{\text{data}}$ and $\mu_{\text{gen}}$ are the means, and $\Sigma_{\text{data}}$ and  $\Sigma_{\text{gen}}$ are the variances.
In practice, we use empirical mean and covariance estimations to compute the FVMD.

%% file: sec/0_figure2.tex
\begin{figure*}[!ht]
    \centering
    \includegraphics[width=\textwidth]{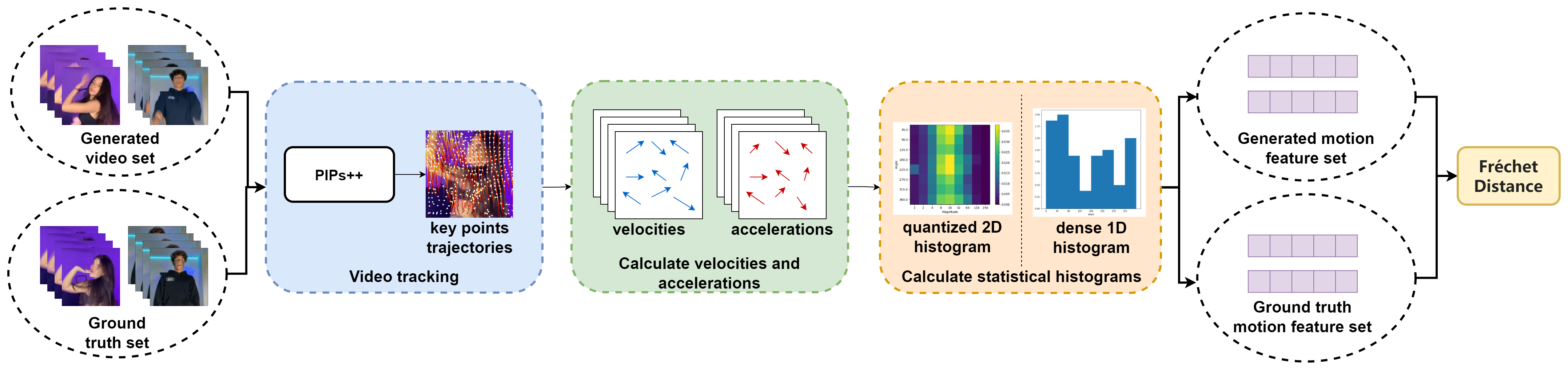}
    \vspace{-0.65cm}
    \caption{
    \textbf{The overall pipeline of our proposed Fr\'echet Video Motion Distance (FVMD).}
    Our pipeline first tracks video key point trajectories using the pre-trained PIPs++~\cite{zheng2023pointodyssey} model and computes the velocity and acceleration fields for each frame. The motion features are then derived from the histograms of the quantized velocity and acceleration. FVMD is eventually given by the Fr\'echet distance between the motion features of generated and ground-truth videos. 
    }
    \label{fig:pipeline}
    \vspace{-0.45cm}
\end{figure*}

%% file: sec/4_experiment.tex
\begin{figure*}[!ht]
    \centering
    \vspace{-0.25cm}
    \includegraphics[width=1.0\textwidth]{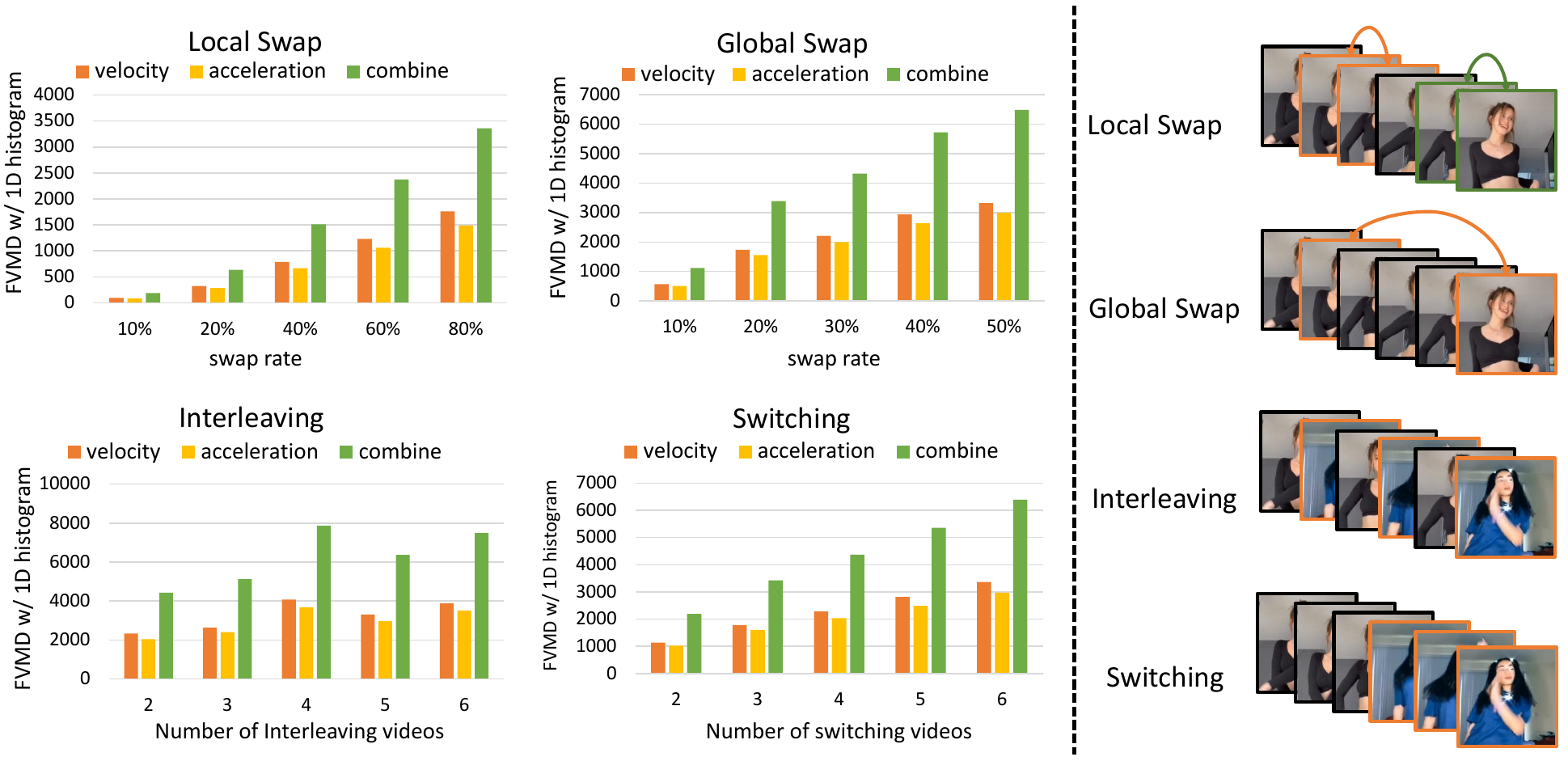} 
    \vspace{-0.6cm}
    \caption{
    \textbf{Sensitivity analysis.}
    We present the FVMD results in the presence of various temporal noises. 
    FVMD based on combined velocity and acceleration features shows the most reliable performance in distinguishing temporal inconsistencies.
    }
    \label{fig:noisy_study}
    \vspace{-1.25em}
\end{figure*}

\begin{table*}[t]
\centering
\resizebox{0.9\textwidth}{!}{
\begin{tabular}{l|cccccc}
\toprule
Metrics & Eql. FVD & Eql. FID-VID & Eql. SSIM & Eql. PSNR & Eql. VBench-AVG & Eql. FVMD \\
\midrule
FVD & - & 0.3596 & 0.0772 & -0.1812 & -0.1898 & -0.7151   \\
FID-VID & -0.1164 & - & 0.3061 & -0.0944 & -0.5226 & -0.6956    \\
SSIM & -0.5926 & -0.7853 & - & -0.8130 & -0.7973 & 0.0527  \\
PSNR & -0.4267 & -0.6096 & 0.1204 & - & -0.7973 & 0.3031  \\
\midrule
VBench-subject consistency & -0.2823 & -0.0386 & 0.262 & -0.2135 & -0.601 & -0.8691 \\
VBench-temporal flickering & -0.5803 & -0.2081 & 0.2413 & -0.128 & -0.4198 & -0.5938 \\
VBench-motion smoothness & -0.4708 & -0.1561 & 0.2698 & -0.0818 & -0.3844 & -0.5448 \\
VBench-dynamic degree & 0.5411 & 0.238 & -0.4063 & 0.0338 & 0.2823 & 0.3894 \\
VBench-imaging quality & 0.5684 & 0.4093 & 0.7160 & 0.8859 & 0.7062 & 0.0383 \\
VBench-AVG & 0.5112 & 0.8835 & 0.341 & 0.2097 & - & -0.5769 \\
\midrule
FVMD & 0.9170 & \textbf{0.9184} & \textbf{0.7191} & \textbf{0.4790} & \textbf{0.7348} & - \\
Combine FVMD \& FVD  & \textbf{0.9173} & 0.8441 & 0.2886 & 0.0383 & 0.4860 & -   \\
\midrule
Agreement rate (human raters) & 0.6773 & 0.8196 & 0.7653 & 0.7184 & 0.7980 & 0.7461   \\
\bottomrule
\end{tabular}
}
\vspace{-0.25cm}
\caption{
\textbf{Pearson correlation for One Metric Equal experiments.} This table shows the Pearson correlation between metrics scores and human perceptions when one selected metric is almost equal, \ie, one can not distinguish these videos relying on the given metric alone. The correlation ranges from $1$ to $-1$, with values closer to $1$ ($-1$) indicating stronger positive (negative) correlation. We also report the agreement rate among raters as a percentage from 0 to 1.
Overall, our FVMD demonstrates the strongest capability to distinguish videos when the other metrics fall short.
}
\vspace{-0.35cm}
\label{tab:equal}
\end{table*}

\section{Experiments}
In~\cref{subsec:sanity}, we conduct a sanity check to verify the soundness of our proposed motion features. In~\cref{subsec:noise}, we conduct sensitivity analysis to demonstrate that our metric is capable of capturing temporal noise. In~\cref{subsec:human}, we carry out large-scale human studies to show that our FVMD is better aligned with human judgment than the existing metrics.
Further, in~\cref{subsec:unary}, we show that our motion features consistently enhance the performance of Video Quality Assessment (VQA) models, suggesting their potential for unary evaluation tasks.
\subsection{Implementation Details}
\label{subsec:details}
We truncate the whole video into segments of $F=16$ frames using a stride of $s=1$ and reshape the spatial size of each frame to $256 \times 256$. We set the number of tracking points to be $N=400$ in all experiments. For the motion feature, we set the small volume shape as $4 \times 5 \times 5$ $ (f=4, k=5)$, so that the quantized 2D histogram feature dimension will be $4 \times 4 \times 4 \times 72$. Similarly, the shape of the dense 1D histogram feature is $4 \times 4 \times 4 \times 8$. 
We empirically identify the velocity-acceleration combined motion feature with a dense 1D histogram as the optimal configuration for our FVMD metric, and thus, it is used as the default. 

\subsection{Sanity Check}
\label{subsec:sanity}
To verify the efficacy of the extracted motion features in representing the motion pattern across a set of videos, we perform a sanity check.
We sample two non-overlapping subsets of videos, randomly drawn from the BAIR video pushing dataset~\cite{ebert2017self}, with different sample sizes. We then evaluate our metrics on these two subsets. 
As claimed in the previous work~\cite{unterthiner2018towards}, the larger the sample size, the better these estimations will be, and the better Fr\'echet distance reflects the true underlying distance between the distributions. As shown in~\cref{{fig:sanity_check}}, our metrics converge to zero as the sample size increases, verifying the hypothesis that the underlying motion distribution within the same dataset should remain consistent.

Furthermore, we extract two subsets of equal sample sizes from two distinct datasets, BAIR video pushing and TikTok dancing~\cite{jafarian2022self}. Our FVMD on these two subsets decreases with the increasing sample size, yet remains higher than the FVMD on two subsets within the same dataset. This observation is in accordance with the assumption that the underlying motion distributions of two different datasets should have a larger gap than the ones within the same dataset. Refer to~\cref{appendix:fvmd_2d} for more sanity check results of FVMD with 2D histogram.

\subsection{Sensitivity Analysis}
\label{subsec:noise}
Following the setting of~\citealp{unterthiner2018towards}, we validate whether our metrics are sensitive to the temporary inconsistency by adding temporal noise to the TikTok dancing dataset~\cite{jafarian2022self}. 
We consider four types of temporary noises: 1) local swap: swapping a fraction of consecutive frames in the video sequence, 2) global swap: swapping a fraction of frames in the video sequence with randomly chosen frames, 3) interleaving: weaving the sequence of frames corresponding to multiple different videos to obtain new videos, 4) switching: jumping from one video to another video.
As shown in~\cref{fig:noisy_study}, our metrics show a strong capability in differentiating various types of injected noise. In particular, the FVMD based on velocity and acceleration combined features has the best performance.
Refer to~\cref{appendix:fvmd_2d} for more results.
\subsection{Human Study}
\label{subsec:human}
An effective video evaluation metric must align with human perceptions. 
We conduct large-scale human studies to validate our proposed FVMD metric.
First, we train a number of conditional video generative models on the TikTok dataset~\cite{jafarian2022self} and draw video samples from their checkpoints. 
We then ask users to compare samples from each pair of models to form a ground-truth user score. 
Subsequently, we calculate the correlation between the score given by each metric and the ground-truth score.

Specifically, we train three different human pose-guided generative models: DisCo~\cite{wang2023disco}, Animate Anyone~\cite{hu2023animate}, and Magic Animate~\cite{xu2023magicanimate}. We fine-tune these models with distinct architectures and hyper-parameters settings, obtaining over 300 checkpoints with different sample qualities. 
We evaluate all checkpoints using our FVMD metric and compare the results with FID-VID~\cite{balaji2019conditional},  FVD~\cite{unterthiner2018towards}, SSIM~\cite{wang2004image}, PSNR~\cite{wang2004image}, and VBench~\cite{huang2023vbench}, which are the most commonly used video evaluation metrics~\cite{melnik2024video}.
For VBench, we evaluate five dimensions related to video quality: subject consistency, temporal flickering, motion smoothness, dynamic degree, and imaging quality. 
Background consistency and aesthetic quality are discarded as they do not support custom videos, and other text prompt-based evaluation dimensions are excluded as inapplicable. We follow the official VBench protocol to calculate an average score for the selected dimensions.

\par\noindent\textbf{Model Selection.}
Following the model selection strategy in~\citealp{unterthiner2018towards}, we design two settings for the human studies. 
The first setup is \textbf{One Metric Equal}.
In this approach, we identify a group of models that have nearly identical scores based on a selected metric. We then investigate whether the other metrics and human raters can effectively differentiate between these models.
Based on the results of the $i$-th metric, we select three groups of model checkpoints corresponding to the quartile points (\ie, top 25\%, 50\%, and 75\%) of its overall distribution, denoted as $\{G_{i,0}, G_{i,1}, G_{i,2}\}$, respectively. Each group contains four models with similar scores for the given metric: $G_{i,j} = \{g_{i,j}^{(0)}, g_{i,j}^{(1)}, g_{i,j}^{(2)}, g_{i,j}^{(3)}\}, j\in \{0,1,2\}$. Each model generates a number of videos, forming groups of video sets denoted as $S_{i,j} = \{S_{i,j}^{(0)}, S_{i,j}^{(1)}, S_{i,j}^{(2)}, S_{i,j}^{(3)}\}$. We then create six pairs of video sets from any two video sets\footnote{The pairs are { $(S_{i,j}^{(0)}, S_{i,j}^{(1)})$, $(S_{i,j}^{(0)}, S_{i,j}^{(2)})$, $(S_{i,j}^{(0)}, S_{i,j}^{(3)})$, $(S_{i,j}^{(1)}, S_{i,j}^{(2)})$, $(S_{i,j}^{(1)}, S_{i,j}^{(3)})$, and $(S_{i,j}^{(2)}, S_{i,j}^{(3)})$.}} for human studies.

The second setting, \textbf{One Metric Diverse}, evaluates the agreement among different metrics when a single metric provides a clear ranking of the performances of the considered video generative models. Specifically, we select model checkpoints whose samples can be clearly differentiated according to the given metric and test the consistency between this metric and other metrics as well as human raters. Similar to the above setups, we select three groups of models, each comprising four checkpoints with significantly different scores for the given metric.
We then draw video samples and construct pairs among them for human studies.

\par\noindent\textbf{Human Rating.}
We ask over 200 individuals to evaluate videos produced by the selected models to study how well the evaluation metrics align with human judgment. For every video set pair, we randomly extract three generated videos. Raters are asked to rate all three video pairs across all model pairs and the most frequently selected option is recorded as the final decision. Following this, we aggregate and determine the user scores for each group by calculating the Borda count~\cite{borda1781m} across all user answers. For more implementation details, please refer to \cref{appendix:human_study_setting}.

\par\noindent\textbf{Evaluation Metrics.}
For each group, we compute the Pearson correlation coefficient between raw scores given by different metrics and the ground-truth human score. Subsequently, the average value across the three groups is computed to represent the final correlation between the metrics and human scores. The higher the value, the better the metric aligns with human judgment.

\begin{table*}[htp!]
\centering
\vspace{-0.15cm}
\resizebox{0.8\textwidth}{!}{
\begin{tabular}{l|cccccc}
\toprule
\multirow{2}{*}{Metrics} & Diverse & Diverse & Diverse & Diverse & Diverse & Diverse \\
                         & FVD & FID-VID & SSIM & PSNR & VBench-AVG & FVMD \\

\midrule
FVD & 0.1007 & 0.1952 & 0.2149 & 0.5662 & -0.1935 & 0.0561   \\
FID-VID & -0.2080 & -0.2002 & 0.0201 & 0.3987 & -0.4441 & -0.0268    \\
SSIM & -0.8617 & -0.5556 & -0.7600 & -0.5515 & -0.6404 & -0.6832 \\
PSNR & -0.6764 & -0.7377 & -0.6812 & -0.6538 & -0.5326 & -0.5842 \\
\midrule
VBench-subject consistency & -0.0102 & -0.3691 &- 0.0452 & 0.1914 & -0.2321 & 0.1819 \\
VBench-temporal flickering & -0.5898 & 0.0755 & -0.0870 & 0.5233 & -0.7701 & -0.5315 \\
VBench-motion smoothness & -0.4563 & 0.1822 & 0.1276 & 0.5936 & -0.6547 & -0.2125 \\
VBench-dynamic degree & \textbf{0.8285} & -0.3992 & 0.2223 & -0.5731 & 0.6866 & 0.7047 \\
VBench-imaging quality & 0.5064 & 0.4593 & \textbf{0.8505} & 0.3655 & 0.6657 & \textbf{0.7404} \\
VBench-AVG & 0.7163 & 0.1720 & 0.5694 & 0.4479 & 0.3031 & 0.4688 \\
\midrule
FVMD & 0.7321 & \textbf{0.8561} & 0.6921 & \textbf{0.9677} & \textbf{0.7928} & 0.6808 \\
Combine FVMD \& FVD  & 0.5940 & 0.5621 & 0.5624 & 0.8192 & 0.5245 & 0.4901   \\
\midrule
Agreement rate (human raters) & 0.8282 & 0.7336 & 0.7529 & 0.7836 & 0.8132 & 0.7665   \\
\bottomrule
\end{tabular}
}
\vspace{-0.25cm}
\caption{
\textbf{Pearson correlation for One Metric Diverse experiments.} This table shows the Pearson correlation between metrics scores and human perceptions when one metric is diverse, \ie, one can distinguish these videos relying on the give metric alone. We also report the agreement rate among raters, which is a percentage ranging from 0 to 1.
}
\vspace{-0.45cm}
\label{tab:diverse}
\end{table*}

\par\noindent\textbf{Results.}
We compare FVMD based on combined velocity-acceleration features and dense 1D histograms with existing metrics, as shown in~\cref{tab:equal} and ~\cref{tab:diverse}. 
Additionally, we explore combining the FVMD with FVD using the F1 score. 
Evidently, our FVMD shows consistently positive and significantly higher correlation coefficients than the other metrics in both the \textbf{One Metric Equal} and \textbf{One Metric Diverse} settings. The quantitative results imply that FVMD is more trustworthy than the baseline metrics.
For more ablation results, please refer to~\cref{appendix:ablation_study}. 
We also report the agreement rate among human raters, which is calculated as the fraction of answers consistent with aggregated answer.
The high agreement among raters indicates their confidence in the survey, enhancing the human study credibility.

In general, the experimental results indicate that our FVMD aligns more closely with human perception across nearly all experimental settings.
In the experiments of \textbf{One Metric Equal}, we observe that FVMD significantly outperforms the other metrics, suggesting that in scenarios where the other metrics fail to evaluate video quality, FVMD can serve as an effective metric to help distinguish videos. 
On the other hand, from the equivalent FVMD column, it is evident that no other metrics can reliably distinguish between models when FVMD results are equal. Moreover, in the experiments of \textbf{One Metric Diverse}, FVMD demonstrates generally higher Pearson correlation than the other metrics. Despite some dimensions of VBench aligning more closely with human perception in certain settings, the overall average score provided by VBench still does not surpass FVMD. Therefore, FVMD is more capable of providing a comprehensive assessment of video quality compared to VBench.

\begin{table}
\centering
\resizebox{\linewidth}{!}{
\begin{tabular}{lcc|cc}
\toprule
\multirow{2}{*}{Method}
& \multicolumn{2}{c}{Vanilla} & \multicolumn{2}{c}{Ours} 
\\ \cmidrule(lr){2-5}
& PLCC $\uparrow$ & SROCC $\uparrow$ & PLCC $\uparrow$ & SROCC $\uparrow$ 
\\ \midrule
 VSFA  & 0.765 & 0.762 & 0.779 & 0.770 \\
 FastVQA & 0.834 & 0.832 & 0.841 & 0.838 \\
 SimpleVQA & 0.847 & 0.840  & {0.870} & {0.861}  \\
\bottomrule
\end{tabular}
}
\vspace{-0.25cm}
\caption{
\textbf{Unary video quality assessment.}
Our motion features consistently boost VQA method performance.
}
\vspace{-.5cm}
\label{tab:unary}
\end{table}

\subsection{Unary Evaluation}
\label{subsec:unary}
FVMD is a pair-wise metric that provides a robust assessment score when a ground-truth video set is available. However, when access to a ground-truth video set is not possible, unary video quality assessment methods become necessary~\cite{liu2024ntire}.
Therefore, we extend the application of our explicit motion features to the Video Quality Assessment (VQA) tasks. We adapt open-source state-of-the-art VQA backbones, including SimpleVQA~\cite{sun2022deep}, FastVQA~\cite{wu2022fast} and VSFA~\cite{li2019quality}, to incorporate our motion feature. We compare their empirical performance on the KVQ dataset~\cite{lu2024kvq}, which is a large-scale VQA benchmark dataset with over 4,000 user-created video clips.
We compare our predicted Mean Opinion Score (MOS) score with the ground-truth score using Pearson linear correlation coefficient (PLCC) and Spearman rank-order correlation coefficients (SROCC). The results are shown in~\cref{tab:unary}. The performances of VQA models are clearly enhanced by our explicit motion features. 

\begin{table}[t]
\centering
\resizebox{\linewidth}{!}{
\begin{tabular}{l|c}
\toprule
Stage & Avg. runtime (sec. per video) \\ 
\midrule
Video tracking &  1.220 \\
Compute vector fields &  0.060 \\
Build 1D histogram & 0.018 \\
Compute Fr\'echet distance & 0.002 \\
\midrule
Overall & 1.325  \\
\midrule
\end{tabular}
}
\vspace{-0.25cm}
\caption{
\textbf{Inference time.}
Most of the runtime is due to video tracking, while other components are light in computation.
}
\vspace{-0.9cm}
\label{tab:runtime}
\end{table}

\subsection{Efficiency}
We report the inference time for each stage of the FVMD pipeline in~\cref{tab:runtime}.
We test the runtime on two subsets of the Tiktok dataset consisting of 1024 16-frame $256\times256$ videos.
The majority of the runtime is consumed by the video tracking stage due to the PIPs++ model.

%% file: sec/5_conclusion.tex
\section{Conclusion}
In this work, we propose a novel metric, \emph{Fr\'echet Video Motion Distance (FVMD)}, to evaluate sample quality for video generative models with a focus on temporal coherence. We design an explicit motion representation based on the patterns of velocity and acceleration in video movements.
Our metric compares the discrepancies of these motion features between the generated and ground-truth video sets, measured by the Fr\'echet distance.
We conduct both sensitivity analysis and human studies to evaluate the effectiveness of our proposed metric.
Our proposed FVMD outperforms existing metrics in many aspects, such as better alignment with human judgment and a stronger capability to distinguish videos of different quality.
Moreover, we validate the promising potential of our motion features for unary video quality assessment through experiments on VQA tasks.

For future directions, we aim to explore a more comprehensive motion representation that conforms to the physical laws of object movement in the real world. This will help detect physically implausible motions and interactions in AI-generated videos, such as abnormal human movements or object trajectories that defy common sense.

%% file: sec/acknowledgement.tex
\section*{Acknowledgements}
This work was funded, in part, by NSERC DG Grants (No. RGPIN-2022-04636 and No. RGPIN-2019-05448), the NSERC Collaborative Research and Development Grant (No. CRDPJ 543676-19), the Vector Institute for AI, Canada CIFAR AI Chair, and Oracle Cloud credits. 
Resources used in preparing this research were provided, in part, by the Province of Ontario, the Government of Canada through the Digital Research Alliance of Canada \url{alliance.can.ca}, and companies sponsoring the Vector Institute \url{www.vectorinstitute.ai/#partners}, Advanced Research Computing at the University of British Columbia, and the Oracle for Research program. 
Additional hardware support was provided by John R. Evans Leaders Fund CFI grant and the Digital Research Alliance of Canada under the Resource Allocation Competition award.

%% file: sec/appendix.tex
\newpage
\appendix
\onecolumn
\section{More Implementation Details}
\subsection{Human study}
\label{appendix:human_study_setting}
For each model group $G_{i,j} = \{g_{i,j}^{(0)}, g_{i,j}^{(1)}, g_{i,j}^{(2)}, g_{i,j}^{(3)}\}$ and its corresponding video sets group $S_{i,j} = \{S_{i,j}^{(0)}, S_{i,j}^{(1)}, S_{i,j}^{(2)}, S_{i,j}^{(3)}\}$, we ask the rater to compare all six generated pairs. For example, the pair $(S_{i,j}^{(0)}, S_{i,j}^{(1)})$, where $S_{i,j}^{(0)}$ and $S_{i,j}^{(1)}$ are sets of videos generated by model $g_{i,j}^{(0)}$ and model $g_{i,j}^{(1)}$ respectively, we randomly select three video pairs that have the same content from them. The rater needs to compare all these three video pairs. If the rater chooses the video generated by model $g_{i,j}^{(0)}$ for two or more of the pairs, then we consider that the rater prefers model $g_{i,j}^{(0)}$. In this case, model $g_{i,j}^{(0)}$ score 1, and model $g_{i,j}^{(1)}$ score 0.

When all raters have completed scoring the six video set pairs, we will sum the scores obtained by models $\{g_{i,j}^{(0)}, g_{i,j}^{(1)}, g_{i,j}^{(2)}, g_{i,j}^{(3)}\}$ respectively to determine the final user score for each model and rank them accordingly

\section{Addition Experiments Results}

\subsection{FVMD with 2D histogram}
\label{appendix:fvmd_2d}
\begin{figure}[h]
    \centering
    \includegraphics[width=0.3\textwidth]{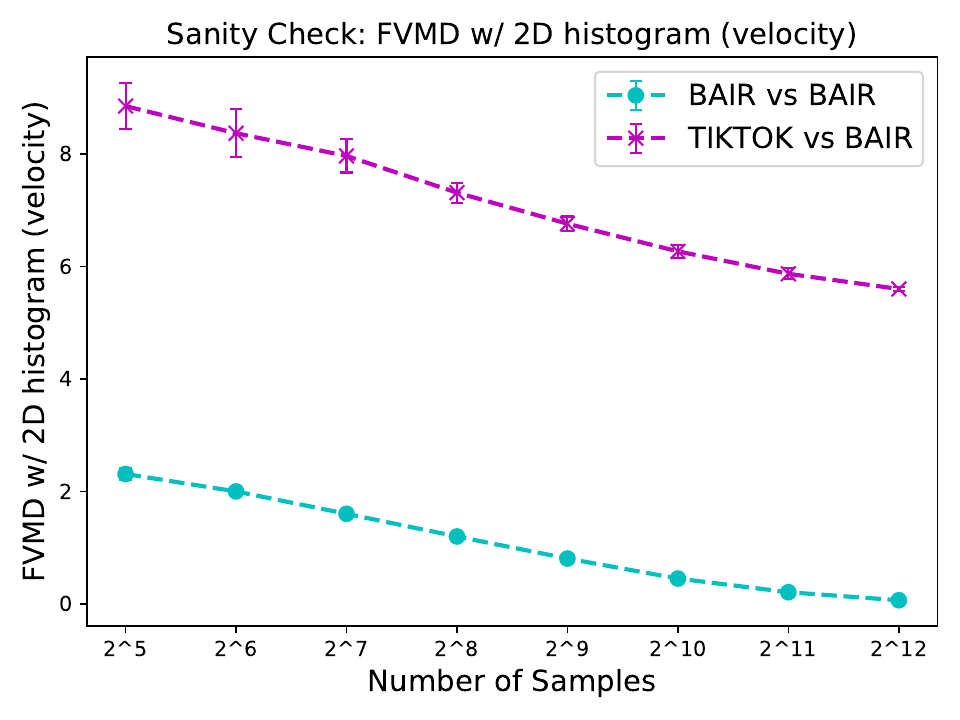} 
    \includegraphics[width=0.3\textwidth]{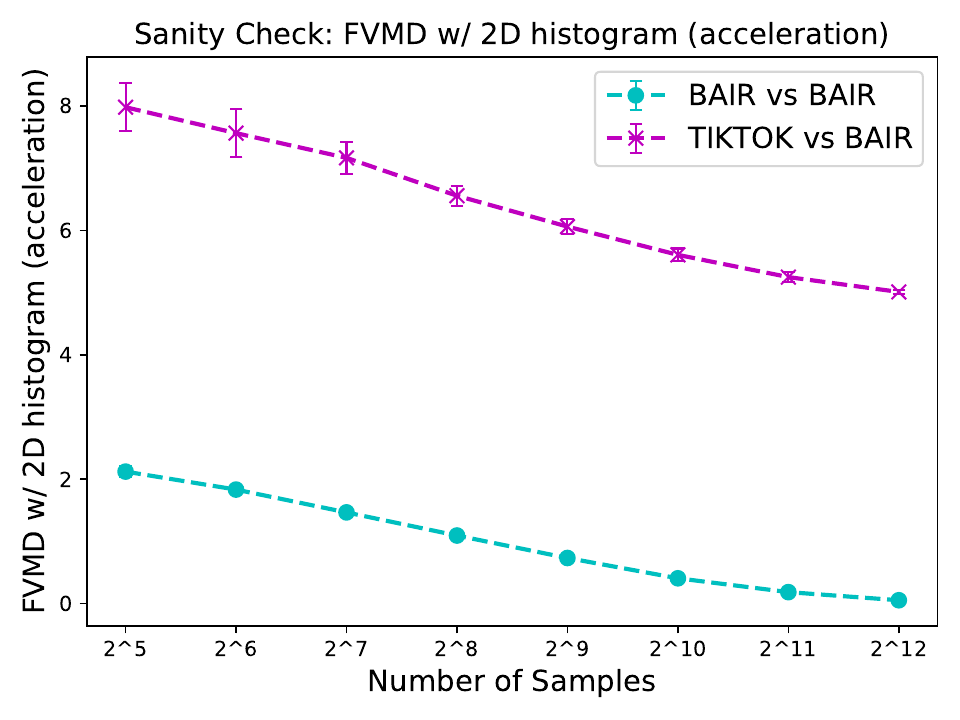} 
    \includegraphics[width=0.3\textwidth]{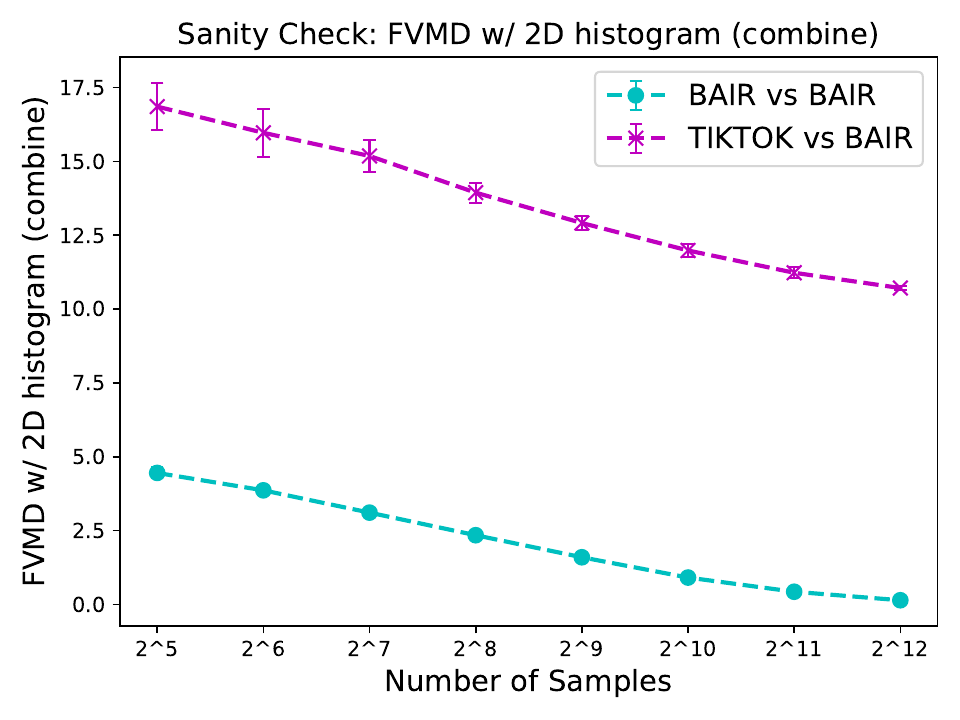} 
    \caption{
    \textbf{Sanity check.} We visualize the curve for FVMD with quantized 2D histogram versus the number of samples. 
    }
    \label{fig:sanity_check_2d}
\end{figure}

\begin{table}[h]
\centering
{
\begin{tabular}{l|c|ccccc}
\toprule
Noise type & Hyperparameter & Int. 1 & Int. 2 & Int. 3 & Int. 4 & Int. 5 \\
\midrule
Local Swap & The proportion of frames swapped & 10\% & 20\% & 40\% & 60\% & 80\% \\
Global Swap & The proportion of frames swapped & 10\% & 20\% & 30\% & 40\% & 50\% \\
Interleaving & The number of videos interleaved & 2 & 3 & 4 & 5 & 6 \\
Switching & The number of videos switched & 2 & 3 & 4 & 5 & 6 \\
\bottomrule
\end{tabular}
}
\caption{
\textbf{Hyperparameter design of the noise study.}
}
\label{tab:noise_hp}
\end{table}

\begin{figure}
    \centering
    \includegraphics[width=1.0\textwidth]{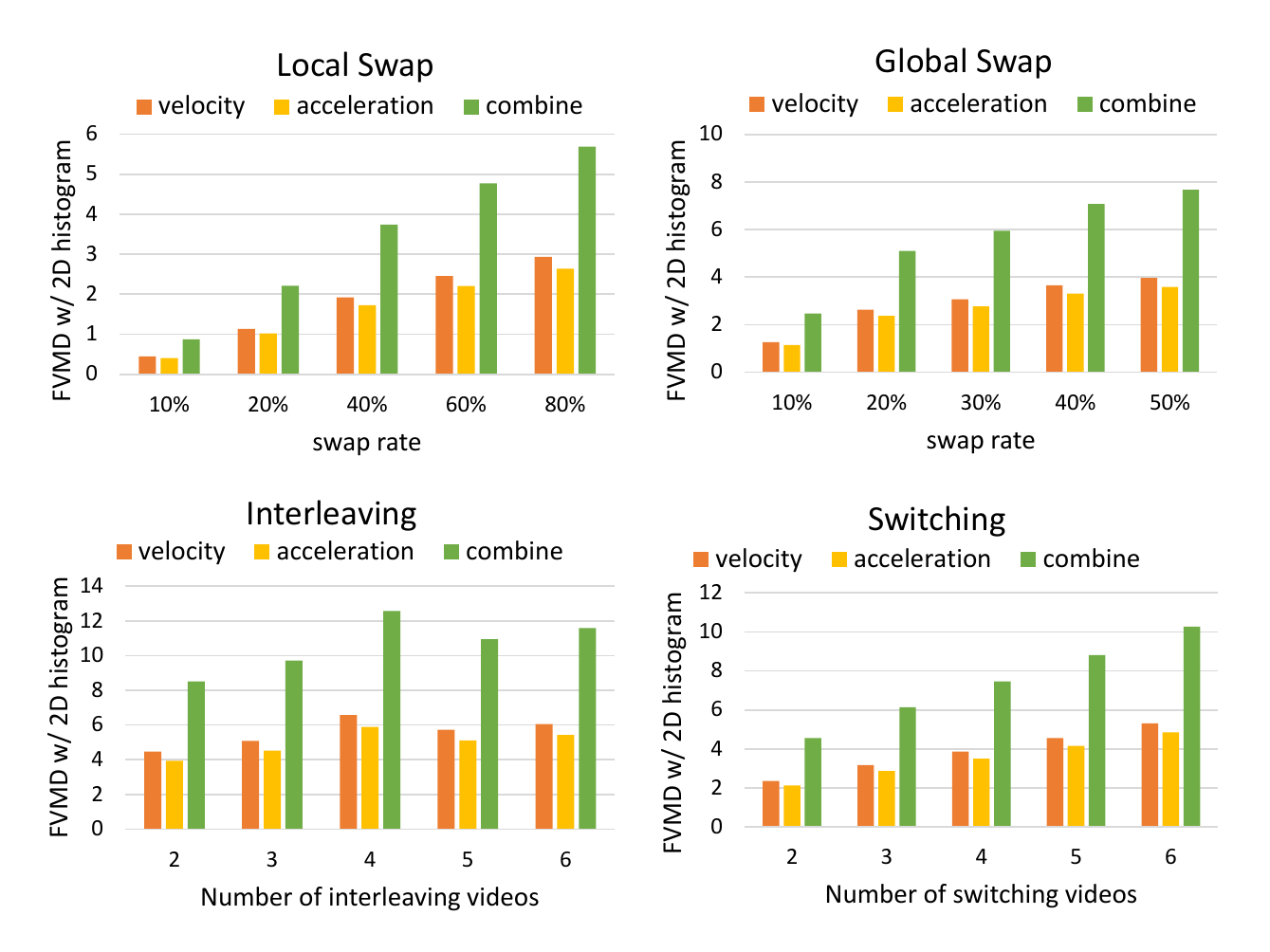} 
    \caption{
    \textbf{Sensitivity analysis}. Behaviors of FVMD with 2D histogram when adding various types of static noise to the temporal dimension of videos. 
    }
    \label{fig:noise_study_2d}
\end{figure}

\textbf{Sanity Check.}
Similar to the sanity check experiments conducted for FVMD with a 1D dense histogram, we also performed a sanity check for the 2D histogram setting. The results, shown in~\cref{fig:sanity_check_2d}, demonstrate that the 2D histogram feature also supports our hypothesis: the underlying motion distribution within the same dataset remains consistent, while the distribution between two different datasets exhibits a larger gap.

\textbf{Sensitivity Analysis.}
We conduct the sensitivity analysis on a subset with a fixed 1024 video clips of the TikTok dataset~\cite{jafarian2022self}. In our sensitivity analysis, the hyperparameter design for the intensity of different types of noise is as shown in~\cref{tab:noise_hp}.
\cref{fig:noise_study_2d} illustrates the behavior of the FVMD with a 2D histogram when various types of static noise are added to the temporal dimension of videos. It is observable that the values of all implementations of FVMD increase with the escalation of added noise.

\clearpage
\subsection{Ablation Study}
\label{appendix:ablation_study}
To determine the optimal configuration for our FVMD, we conduct ablation experiments under the same experimental setup as used in the human study. We explore alternative designs for the motion features, including: 1) different motion representations, including computing only velocity fields, only acceleration fields, and combining velocity and acceleration; 2) different methods for statistically characterizing vector fields, including quantized 2D histograms and dense 1D histograms; 3) the degree of overlap when extracting 16-frame segments from the entire video, ranging from no overlap (stride=16) to maximum overlap (stride=1). The results of the ablation study are presented in~\cref{tab:ablation_eql} and~\cref{tab:ablation_divers}.

It is evident that different motion representations do not significantly impact the performance of our metric. Additionally, the performance of FVMD with a dense 1D histogram surpasses that of FVMD with a quantized 2D histogram. For FVMD with a dense 1D histogram, the maximum overlap strategy when extracting video clips outperforms the non-overlap strategy across all experimental setups. Overall, FVMD utilizing a combined motion representation with a dense 1D histogram and maximum overlap video segments aligns more closely with human perception. Therefore, we select this as the default configuration for our FVMD metric.

\begin{table}
\centering
\resizebox{\textwidth}{!}{
\begin{tabular}{ccc|cccccc}
\toprule
\multicolumn{3}{c|}{Configuration} & \multirow{2}{*}{eql. FVD} & \multirow{2}{*}{eql. FID-VID} & \multirow{2}{*}{eql. SSIM} & \multirow{2}{*}{eql. PSNR} & \multirow{2}{*}{eql. VBench-overall score} \\ \cmidrule(lr){1-3}
motion representation & statistics & stride $s$ \\
\midrule
velocity & 2D & 16 & 0.9314 & 0.8469 & 0.4263 & 0.0448 & 0.0117    \\
acceleration & 2D & 16 & \textbf{0.9316} & 0.8453 & 0.4216 & 0.0362 & -0.0063     \\
combine & 2D & 16 & 0.9315 & 0.8464 & 0.4241 & 0.0423 & 0.0039   \\
\midrule
velocity & 1D & 16 & 0.8773 & 0.9210 & 0.4934 & 0.2007 & 0.5642   \\
acceleration & 1D & 16 & 0.8601 & 0.8985 & 0.5104 & 0.1993 & 0.5510  \\
combine & 1D & 16 & 0.8612 & 0.9091 & 0.4925 & 0.1894 & 0.5414   \\
\midrule
velocity & 2D & 1 & 0.8555 & 0.8106 & 0.3903 & 0.0080  & -0.7144  \\
acceleration & 2D & 1 & 0.8627 & 0.8136 & 0.3913 & 0.0115 & -0.1101  \\
combine & 2D & 1 & 0.8569 & 0.8115 & 0.3916 & 0.0109 & -0.1144   \\
\midrule
velocity & 1D & 1 & 0.9172 & \textbf{0.9253} & 0.7128 & \textbf{0.4920} & \textbf{0.7359}  \\
acceleration & 1D & 1 & 0.9276 & 0.9112 & 0.7162 & 0.4851 & 0.7354 \\
combine & 1D & 1 & 0.9170 & 0.9184 & \textbf{0.7191} & 0.479 & 0.7348  \\
\bottomrule
\end{tabular}
}
\caption{
\textbf{Ablation study on One Metric Equal setting.} The experimental setup is consistent with that described in~\cref{tab:equal}. The eql. FMD column has been omitted.
}
\label{tab:ablation_eql}
\end{table}

\begin{table}
\centering
\resizebox{\textwidth}{!}{
\begin{tabular}{ccc|cccccc}
\toprule
\multicolumn{3}{c|}{Configuration} & \multirow{2}{*}{divers. FVD} & \multirow{2}{*}{divers. FID-VID} & \multirow{2}{*}{divers. SSIM} & \multirow{2}{*}{divers. PSNR} & \multirow{2}{*}{divers. VBench-overall score} & \multirow{2}{*}{divers. FVMD} \\ \cmidrule(lr){1-3}
motion representation & statistics & stride $s$ \\
\midrule
velocity & 2D & 16 & 0.5851 & 0.2831 & 0.6209 & 0.6977 & 0.3275 & 0.3985   \\
acceleration & 2D & 16 & 0.5791 & 0.2893 & 0.6136 & 0.7013 & 0.3169 & 0.3936   \\
combine & 2D & 16 & 0.5838 & 0.2880 & 0.6189 & 0.7008 & 0.3232 & 0.3963  \\
\midrule
velocity & 1D & 16 & 0.6365 & 0.6920 & 0.6126 & 0.8866 & 0.6835 & 0.5768  \\
acceleration & 1D & 16 & 0.6282 & 0.7016 & 0.6100 & 0.8929 & 0.6854 & 0.5750 \\
combine & 1D & 16 & 0.6269 & 0.6910 & 0.6085 & 0.8866 & 0.6781 & 0.5699  \\
\midrule
velocity & 2D & 1 & 0.5075 & 0.2714 & 0.4803 & 0.7162 & 0.3358 & 0.3942  \\
acceleration & 2D & 1 & 0.5076 & 0.2776 & 0.4820 & 0.7196 & 0.3424 & 0.3946 \\
combine & 2D & 1 & 0.5082 & 0.2772 & 0.4823 & 0.7193 & 0.3399 & 0.3945  \\
\midrule
velocity & 1D & 1 & \textbf{0.7388} & \textbf{0.8588} & \textbf{0.6959} & \textbf{0.9685} & 0.7951 & \textbf{0.6836}  \\
acceleration & 1D & 1 & 0.7311 & 0.8582 & 0.6905 & 0.9665 & \textbf{0.7952} & 0.6835 \\
combine & 1D & 1 & 0.7321 & 0.8561 & 0.6921 & 0.9677 & 0.7928 & 0.6808  \\
\bottomrule
\end{tabular}
}
\caption{
\textbf{Ablation study on One Metric Diverse setting} The experimental setup is consistent with that described in~\cref{tab:diverse}.
}
\label{tab:ablation_divers}
\end{table}